\documentclass[conference]{IEEEtran}
\usepackage{latexsym}
\usepackage{adjustbox}
\usepackage{multicol}
\usepackage{hyperref}
\usepackage{afterpage}
\usepackage{cite}
\usepackage{amsmath,amssymb,amsfonts, bbm, array}
\usepackage{algorithmic}
\usepackage{diagbox}
\usepackage{multirow}
\usepackage{makecell}
\usepackage{graphicx}
\usepackage{url}
\usepackage{textcomp}
\usepackage{xcolor}
\def\BibTeX{{\rm B\kern-.05em{\sc i\kern-.025em b}\kern-.08em
    T\kern-.1667em\lower.7ex\hbox{E}\kern-.125emX}}
\begin{document}

\title{
From Generalized Laughter to Personalized Chuckles:
Unleashing the Power of Data Fusion in Subjective Humor Detection
}

\author{\IEEEauthorblockN{
Julita Bielaniewicz and
Przemysław Kazienko}
\IEEEauthorblockA{\textit{Department of Artificial Intelligence, Wrocław University of Science and Technology, Poland}}
\IEEEauthorblockA{\footnotesize{\texttt{\{julita.bielaniewicz, kazienko\}@pwr.edu.pl}}} 
}


\maketitle

\begin{abstract}
The vast area of subjectivity in Natural Language Processing (NLP) poses a challenge to the solutions typically used in generalized tasks. As exploration in the scope of generalized NLP is much more advanced, it implies the tremendous gap that is still to be addressed amongst all feasible tasks where an opinion, taste, or feelings are inherent, thus creating a need for a solution, where a data fusion could take place. We have chosen the task of funniness, as it heavily relies on the sense of humor, which is fundamentally subjective. 
Our experiments across five personalized and four generalized datasets involving several personalized deep neural architectures have shown that the task of humor detection greatly benefits from the inclusion of personalized data in the training process. We tested five scenarios of training data fusion that focused on either generalized (majority voting) or personalized approaches to humor detection. The best results were obtained for the setup, in which all available personalized datasets were joined to train the personalized reasoning model. It boosted the prediction performance by up to approximately 35\% of the macro F1 score. Such a significant gain was observed for all five personalized test sets. At the same time, the impact of the model's architecture was much less than the personalization itself. 
It seems that concatenating personalized datasets, even with the cost of normalizing the range of annotations across all datasets, if combined with the personalized models, results in an enormous increase in the performance of humor detection.
\end{abstract}


\section{Introduction}
\label{sec:introduction}
Recent years have shown that the subjective field of natural language processing is gradually receiving more attention, yet still comparatively less than the predominant generalization approach. This is especially apparent when focusing on a single personalization task, such as emotion detection, hate speech, or funniness. Although this implicitly signals the difference in the amount of research, the possibilities that lie in personalized data utilization even in a single subjective task are very promising\cite{kanclerz2020cross, kanclerz2021controversy}.
An indirect issue that tends to be ignored is the nature of groups and subgroups that can be distinguished according to general characteristics in the sense of humor of individuals. Those groups could be sorted according to age, gender, political views, occupation, and many other characteristics \cite{hofmann2020gender}. Undoubtedly, this will allow for a more accurate image of the sense of humor per group, but will still remain overall generalized. What is crucial is that in each group we would still observe unique individuals who end up omitted from the majority. Regardless, following this path, we could create subgroups in the said groups that will sort users more accurately, yet still have a generalization approach. Furthermore, this would lead to more and more subgroups that will set the gold standard of a single group anew, still bearing the generalization, until we focus on each person separately. This final product sets a standard for the sense of humor per person, which is, in essence, humor personalization. In this way, we eliminate the possibility that anyone with an unusual sense of humor would be ignored. 
Given the prospect concealed within personalization \cite{kocon2021learning, kocon2023chatgpt, mieleszczenko2023capturing, kazienko2023human}, there was a need for scenarios where there was the possibility of checking different combinations of subjectivity and its impact on the model performance. We intended to implement data fusion that could improve the overall performance of personalized reasoning by showing which combination of subjectivity to generalization ratio was optimal. Our research resulted in discovering the major gain for the models trained on a fully subjective fused dataset, as well as noticing that it effectively increases the quality of personalized deep learning architectures. This achievement is especially exciting, as it opens the door to the possibilities that lie within the data fusion domain.

In this work, our aim is to answer the following research questions:
\begin{enumerate}
\item Does providing knowledge from other datasets help the model better understand the task of humor detection? (see Sec.~\ref{sec:single_vs_multi})
\item Does the fusion of data about individual users sense of humor from various datasets affect the reasoning quality for other personalized datasets in personalized and majority voting-based scenarios? (see Sec.~\ref{sec:single_vs_multi})
\item Does the addition of generalized datasets improve the quality of humor prediction for personalized datasets in a majority voting scenario? (see Sec.~\ref{sec:knowledge_transfer_majority_personalization})
\item Does the knowledge transfer between personalized datasets affect the model performance in a similar way as the knowledge transfer between the datasets with majority voting? (see Sec.~\ref{sec:knowledge_transfer_majority_personalization})
\item Does the gain received from data fusion depend on the language or the domain of the datasets? (see Sec.~\ref{sec:knowledge_transfer_majority_personalization})
\item Does the use of personalized data fusion techniques improve the prediction quality of various personalized architectures in a similar way?  (see Sec.~\ref{sec:data_fusion_on_architectures})
\end {enumerate}

\section{Related work}
\label{sec:related_work}
There is no doubt that tasks that are subjective in nature in the area of Natural Language Processing tend to receive a somewhat dismissive treatment, being simplified in order to analyze what kind of content the majority of users are anticipating. Despite best efforts, this setup divides all users into people who fall into the general majority of each subset, who are directly involved throughout all research, and users who end up being abruptly cut off from the analysis right from the very beginning. That group of people never had the opportunity to be included because of their diversity compared to their respectful groups. This way, a huge part of useful information is lost, simply because the initial assumption is based on generalization. This simplification relies on the absence of user information and, thus, is made up of texts and annotations. After all, the sole detection of an anticipated feature does not need data on the subjective perception of a given content as such. However, we argue that setting the gold standard per user enriches the knowledge about the chosen area of NLP.
The usual tendency in the scope of Natural Language Processing when aiming for better results in generalized humor detection is to create and utilize many different categorizations.\cite{wilson1969conservatism, bekelja2006appropriate} The most prominent, 12 types of humor included in the Theory of Humor \cite{hay1995gender} have withstanded the test of time in the span of the last 30 years and remain relevant in humor detection to this day, setting the ground for many humor-related research \cite{strapparava2011computational, raz2012automatic, khandelwal2018humor} as well as other categorizations of humor in text \cite{dynel2009beyond, davies2017sociolinguistic}.
This research provides a general insight into why some people may find something humorous while others may not. The analysis in this scope is focused on standard per group. It was determined to be a key factor in the realm of user-centric funniness if outliers of humor were omitted \cite{agrawal2020joke}. Comparing different jokes can help identify groups of people who have a similar sense of humor. Calculating the similarity between different jokes facilitates identifying human groups with a common sense of humor. Nonetheless, more defined groups are only scratching the surface of possibilities that can be easily uncovered if switching to a personalized approach. If a goal towards capturing a group of users accurately enough to hit the target includes outliers who may grow indifferent to such a solution, it is not viable in the long run in our reality.
Humor is a subjective experience that can bring about varying degrees of amusement in its viewers, but can also lead to other unexpected responses, such as feeling insulted \cite{meghana2020humour, siddiqui2019sarcasmania}. If analyzing, for example, dark humor, many people are not enjoying it purely because of different levels of tolerance and sensitivity to disturbing, sometimes also offensive humor. As such, it may not be an extreme situation, yet it is possible that a user in certain humor groups will receive content upsetting enough to experience mental turmoil, only because of their age group or other categorizing factors. This experience is completely opposite to planned, despite best efforts, because the humor norm was simplified. 
With this in mind, it is better to use personalization when possible. If these conditions are met, and therefore we have access to different personalization datasets, it is time to strategize data fusion \cite{bleiholder2009data} to obtain an effective prediction quality. Research on the specific scope of humor data fusion does not focus on personalization \cite{xu2022hybrid, christ2022multimodal, christ2023muse} and so we propose our own take on this domain.

\section{Data Fusion in Subjective Humor Detection}
\label{sec:data_fusion}

\subsection{Personalized Humor Detection}
\label{sec:humor_personalization}

\begin{figure}[h]
    \centering
    \includegraphics[width=\linewidth]{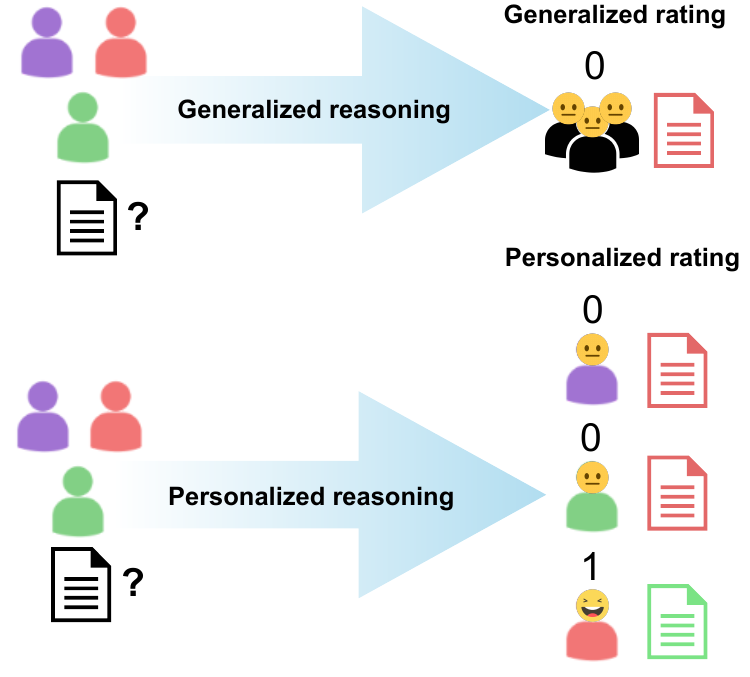}
    \caption{Difference between the generalized (top) and personalized (bottom) perspective. In the first one, the same prediction is provided for every user. In the second one, the model outputs various predictions on the basis of individual user preferences.
    }
    \label{fig:generalized_vs_personalized}
\end{figure}

The feeling of amusement is experienced differently by each person. This phenomenon is further enhanced in the case of texts, where the choice of words and their diverse impact on a particular user also play an important role. Therefore, the natural need to model the individual user's sense of humor is outlined. To quantify the funniness of a text, a binary classification\cite{yang2015, castro2018, humicroedit_dataset, hossain-etal-2020-semeval} is often used: \textit{funny} (class 1) or \textit{not funny} (class 0). This generalized approach ignores the peculiarities of human perspective and assumes training the model to predict the same class for every user (see Fig.~\ref{fig:generalized_vs_personalized}, top).

The idea of personalized humor detection\cite{bielaniewicz2022} is quite different. It uses the assumption that the individual user's sense of humor should be taken into account when predicting the funniness of a text along with the content of the evaluated text. In this approach, the trained model provides various predictions for different people. The personalized humor detection approach is presented in the bottom row of Fig.~\ref{fig:generalized_vs_personalized}.

\subsection{Human-Centered Data Fusion}
\label{sec:personalized_data_fusion}

\begin{figure}[h]
    \centering
    \includegraphics[width=\linewidth]{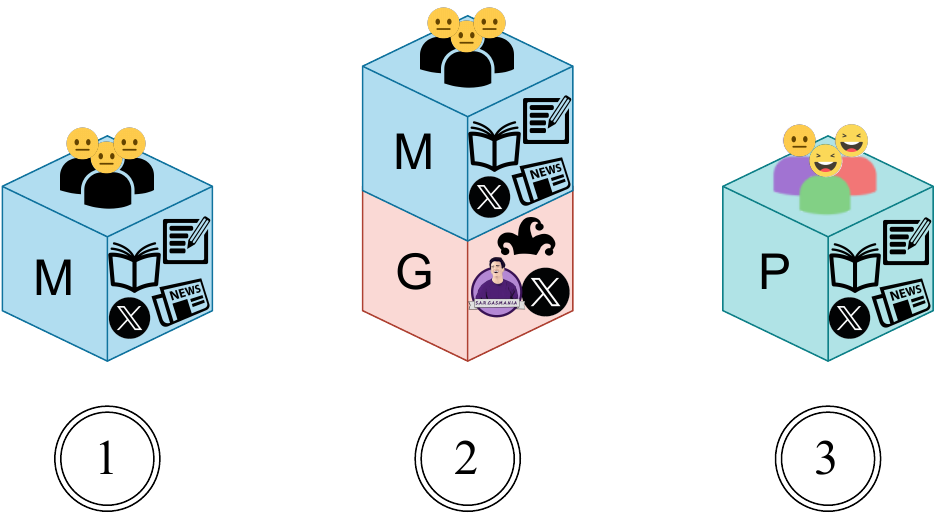}
    \caption{Our human-centered data fusion approaches: (1) multiple combined majority voting datasets, (2) majority voting datasets combined with generalized datasets, and (3) multiple personalized datasets.
    }
    \label{fig:data_fusion}
\end{figure}

The perception of humor in a text is a personality trait characterized by a very high subjectivity\cite{ziv1979sociometry}. To tackle this problem, we propose a data fusion approach focused on combining user annotations from various humor datasets. The basic variant assumes the integration of knowledge from multiple humor datasets with user annotations aggregated via majority voting to reduce the uncertainty caused by multiple users annotating texts from various domains and written in diverse languages (first column in Fig.~\ref{fig:data_fusion}). Our technique can also be used to synthesize the knowledge obtained from the majority voting annotations with generic knowledge about the funniness of textual content from generalized datasets (second column in Fig.~\ref{fig:data_fusion}). The most advanced variant of our method leverages the original user annotations to preserve every person's individual sense of humor. In this way, we maximize the available knowledge that the model can acquire during the training procedure (third column in Fig.~\ref{fig:data_fusion}).

\section{Datasets}
\label{sec:datasets}
We have chosen 9 datasets that consisted of annotated pieces of humorous text in the form of individual words, phrases, or even pairs of text before and after a slight change. 5 of those datasets were strictly personalized, which means that they included information about the user IDs, as well as many annotations for an independent text.
The remaining 4 datasets are generalized, with only one annotation per text available. The vast majority of texts were in English, but there are also ones in Spanish, as well as Polish languages. More detailed information on personalized datasets is available in Tab.~\ref{tab:datasets_per} and the generalized datasets are described in Tab.~\ref{tab:datasets_gen}
\subsection{Personalized datasets}
\label{sec:personalized_datasets}
\subsubsection{Cockamamie Gobbledegook}
This dataset \cite{engelthaler2018humor} is made of 10,000 English word expressions (1-2 words), usually word formations. Annotators are workers of the Amazon Mechanical Turk of the United States. For experiment purposes, we used only words that had at least one positive and one negative annotation in order to better distinguish between user preferences. 
\subsubsection {Humor} A dataset \cite{castro2018} consisting of 27k Spanish tweets annotated by crowd-sourcing workers. 
We have used the version\footnote{\url{https://github.com/pln-fing-udelar/humor/tree/main/previous}} of the dataset that contains text, not only tweet IDs. For the purposes of our experiments, we have considered only texts with more than one annotation. 
\subsubsection{Humicroedit} \cite{humicroedit_dataset} was used for competition in the computational recognition of humor in SemEval-2020 Task 7 \cite{hossain-etal-2020-semeval}. The authors used Amazon Mechanical Turk annotators to edit a single word of 14k news headlines to make them funny. 
The embeddings of both original and edited texts with their scores are provided at the input to all models so that the context is taken into account in humor recognition.
\subsubsection{Doccano 1}
This dataset is one of the iterations of the Doccano 1.0 project, which captures the perceptions and feelings evoked by textual content.
Each text has a length of up to 132 words ($\mu$ = 24.5, $\sigma$ = 16.2). The average number of texts annotated by each person was approximately 790, and the average number of annotators for each text was approximately 32. In total, the number of annotations is around 31,700.
On average, labels with zero value appear 62\% of the time, with a standard deviation of 22\%. Additionally, empty labels are seen 4\% of the time, with a standard deviation of 8\%.
\subsubsection{Doccano 2}
The dataset which, similarly to Doccano 1, is one of the iterations of the Doccano 1.0 project. 
On average, each person annotated around 358 texts, and each was annotated by 2 annotators. In its entirety, it is a little under 17,700 annotations. As in Doccano 1, each annotation consists of 26 independent dimensions.
The possible annotation values were set to be chosen during the annotation procedure in the same way as in Doccano 1.
\subsection{Generalized datasets}
\label{sec:generalized_datasets}
\setcounter{subsubsection}{5}
    \subsubsection{ColBERT} a large dataset for the task of humor detection. Several existing humor detection datasets combined some non-humorous formal texts with short informal humorous texts.\cite{annamoradnejad2020colbert} This dataset contains 200,000 short texts (100,000 positive and 100,000 negative).
    \subsubsection{HahaIberlef2021} is a corpus of crowd-annotated tweets separated into three subsets: training (24,000 tweets), development (6,000 tweets), and testing (6,000 tweets) \cite{chiruzzo2021overview}. The annotation uses a voting scheme in which users could select one of six options \cite{chiruzzo2020haha}: the tweet is not humorous, or the tweet is humorous, and a score is given between one (not funny) and five (excellent).
    \subsubsection{Hahackathon} a dataset that contains 10,000 humor and offense annotated texts by 20 annotators aged 18 to 70 years. \cite{meaney-etal-2021-semeval} It had 80\% of its data sourced from Twitter. The remaining 20\% of the texts were selected from the Kaggle Short Jokes dataset.
    \subsubsection{Sarcasmania} is a Kaggle-sourced dataset that consists of 39,780 texts from the Twitter platform. \cite{siddiqui2019sarcasmania} Each text is annotated for three dimensions: "sarcasm", "humor", and "insult". To incorporate this dataset, we have focused purely on the humor aspect of annotations within this collection of texts.

\begin{table*}
\centering
\caption{Personalized dataset profiles. Each dataset contains a set number of labels. Details and further information about them can be found in Section \ref{sec:datasets}.}
\begin{tabular}{|l|c|c|c|c|c|}
\hline
\diagbox[width=15em]{Property}{Dataset} & \makecell{Cockamamie \\ Gobbledegook} & Humor & Humicroedit & Doccano 1 & Doccano 2  \\
\hline
Textual content profile & 1-2 words & tweets & paired news \& headlines & comments & comments \\
\hline
No. of texts & 10,884 & 8,284 & 14,886 & 880 & 8,891 \\
\hline
No. of annotations &  40,673  &  26,967 & 74,430 & 31,521 & 17,533 \\
\hline
Number of annotators & 351 &  3,137 & 5 & 39 & 49 \\
\hline
Avg. annotations per text  & 3.74 & 3.26 & 5 & 35.81 & 1.97 \\
\hline
Avg. annotations per annotator & 115.88 & 57.44 & 14,886 & 808.23 & 357.81 \\
\hline
Class balance (0/1) & 382,083 / 51,197 & 12,091 / 9,978 & 31,863 / 42,572 & 26,706 / 4,815 & 7,661 / 9,872 \\
\hline
Language & English & Spanish & English & Polish & Polish \\
\hline
\end{tabular}
\label{tab:datasets_per}
\end{table*}

\begin{table*}
\centering
\caption{Generalized dataset profiles. Each dataset contains a set number of labels. The number of annotations is equal to the number of tests because of the lack of any user distinction. Details and further information about them can be found in Section \ref{sec:datasets}.
}
\begin{tabular}{|l|c|c|c|c|c|}
\hline
\diagbox[width=15em]{Property}{Dataset} & ColBERT & HahaIberlef2021 & Hahackathon & Sarcasmania \\
\hline
Textual content profile & short texts & tweets & tweets \& jokes & tweets \\
\hline
No. of texts & 199,996 & 33,824 & 7,958 & 39,778 \\
\hline
Class balance (0/1) & 100,000 / 99,996 & 20,778 / 13,046 & 3,068 / 4,890 & 20,135 / 19,643 \\
\hline
Language & English & Spanish & English & English
\\
\hline
\end{tabular}
\label{tab:datasets_gen}
\end{table*}

\section{Models}
In the personalized approach, it is assumed that the funniness level of a text is determined by the users and their individual sense of humor, rather than just the text itself and the average or combined annotations. This premise is based on the user's individual preferences, which should be utilized in order to receive a distinct output for each user, based on their individual preferences as presented in Fig.~\ref{fig:generalized_vs_personalized}. For this purpose, we need the information about user preferences extracted from their past annotations, and the models in \cite{bielaniewicz2022} are doing exactly that, which is why we use them in our research. As an addition to the deep learning models, we also leveraged a slightly more complex model, the UserId architecture. It was sourced from the work \cite{kocon2021ipm}
\label{sec:models}

\subsection{TXT-Baseline} This deep neural architecture \cite{bielaniewicz2022} has an input of only a text embedding.It consists of the layers of the deep language model and the layers associated with the learning task. It is characterized by representing a generalized method with a unified output for all users, widely used in NLP. The main goal of exploiting this model is to accurately measure and compare differences between generalized and personalized approaches.

\subsection{OneHot} Consisting of a most simple and yet very effective method for the representation of annotators, this model \cite{bielaniewicz2022} utilizes information about the user ID to incorporate it into a one-hot vector. A simple variant of personalized architectures.

\subsection{SHEEP-Formula} This architecture \cite{bielaniewicz2022} is characterized by the use of a metric called Human Sense of Humor, i.e. a \textit{HSH(u)}. This value is sent directly to the model input, and a single estimate is calculated for each user individually and has its basis on the Z-score.
When the SHEEP-Formula clashes with a random language model, the learning and reasoning processes are based solely on the information about the user.

\subsection{SHEEP-Simple} SHEEP-Simple \cite{bielaniewicz2022} is a straightforward personalized model, which builds upon TXT-Baseline's textual prediction by incorporating the individual's preferences and the words used in the text. Biases are incorporated into the final prediction, which can be interpreted as adjusting the inference for a particular individual by taking into account their average annotation in the dataset. The biases are set randomly at the start and then learned through backpropagation. This model focuses on neither the relation between a user nor the text, but only the statistics of human annotations.

\subsection{SHEEP-Medium} This model \cite{bielaniewicz2022} trains a multidimensional latent vector of an annotator to grasp the knowledge of their sense of humor.
To address the problem of performance in personalization during the processing of unknown texts during model training, SHEEP-Medium utilizes a vector representation of a human. This allows for capturing their approach to a specific text in the context of their sense of humor.

\subsection{UserId} The model architecture \cite{kocon2021ipm} is based on appending the user ID token to the start of the annotated text in order to store information about a person. Subsequently, the text containing the user ID is transformed into a vector representation using a transformer model. To prevent the tokenizer from splitting the user ID tokens,  the original model was extended by manually adding them to the special tokens set of the model. The transformer weights were trained with the entire model to learn the dependencies between the user and the text.

\section{Experimental setup}
\label{sec:experimental_setup}
To evaluate our human-centered data fusion techniques we leveraged several experimental configurations: (1) training the model on the samples from a single dataset aggregated via majority voting (\textbf{Majority single}), (2) combining samples from multiple personalized datasets aggregated via majority voting in the train set (the first column in Fig.~\ref{fig:data_fusion}, \textbf{Majority multi}), (3) fusing samples from multiple personalized datasets and generalized datasets during model training (the middle column in Fig.~\ref{fig:data_fusion}, \textbf{Majority + Generalized multi}), (4) training the model on individual user annotations from a single personalized dataset (\textbf{Personalized single}), (5) combining individual user annotations from all personalized datasets in the train set (the last, third column in Fig.~\ref{fig:data_fusion}, \textbf{Personalized multi}).
To enable the use of our data fusion techniques for any of the selected datasets regardless of their diverse label ranges, we mapped any value different than zero to class 1 (\textit{funny}). In this way, we preserved the initial information about the funniness of the text, which was also the source of knowledge about the user perspective in the case of personalized datasets presented in Sec.~\ref{sec:personalized_datasets}.

\begin{figure}[h]
    \centering
    \includegraphics[width=\linewidth]{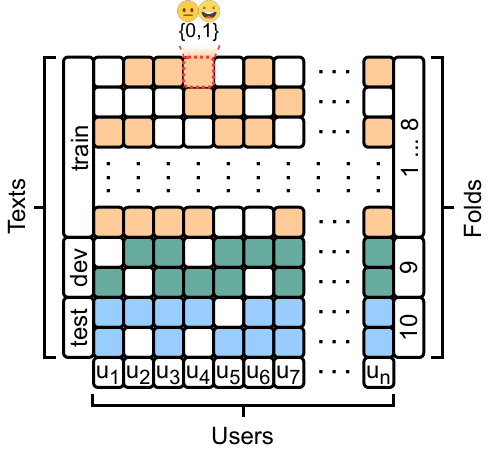}
    \caption{Data split strategy used for each dataset. White blocks are texts, which are not annoated by a specific user}
    \label{fig:data_split}
\end{figure}

We addressed any potential disparity and data leak between text samples by utilizing the text-based data division described in Fig.~\ref{fig:data_split}. In this setup, the model is validated and tested only on texts that were not present in the train set.

To measure the stability of the results, we used the 10-fold cross-validation. During each iteration, the model was trained on 8 folds, while {9\textsuperscript{th}} and {10\textsuperscript{th}} fold were used as validation and test sets, respectively. In the next step, we calculated the mean and standard deviation of the selected evaluation metrics. Furthermore, we measured the statistical significance of the obtained results via statistical tests. For data that satisfied the test assumptions, we used the \textit{t}-test for independent samples \cite{student1908} with Bonferroni correction \cite{bonferroni1936teoria}. Otherwise, we leveraged the Mann-Whitney \textit{U}-test \cite{mann1947}.

For evaluation, we used the macro F1 score \cite{sokolova2006beyond} and the \textit{Gain} measure, which is equal to the difference between the macro F1 score achieved by the model trained on the dataset obtained via data fusion and the macro F1 score of the model trained on the original dataset. 

Due to the multilingual character of the datasets included in the experiments, we obtained the text representations through the LaBSE \cite{feng-etal-2022-language} model available in the HuggingFace \textit{transformers} library \cite{wolf-etal-2020-transformers}.

\section{Results}
\label{sec:results}

\subsection{Evaluation of Data Fusion Techniques}
\label{sec:single_vs_multi}

\begin{figure}[h]
    \centering
    \includegraphics[width=\linewidth]{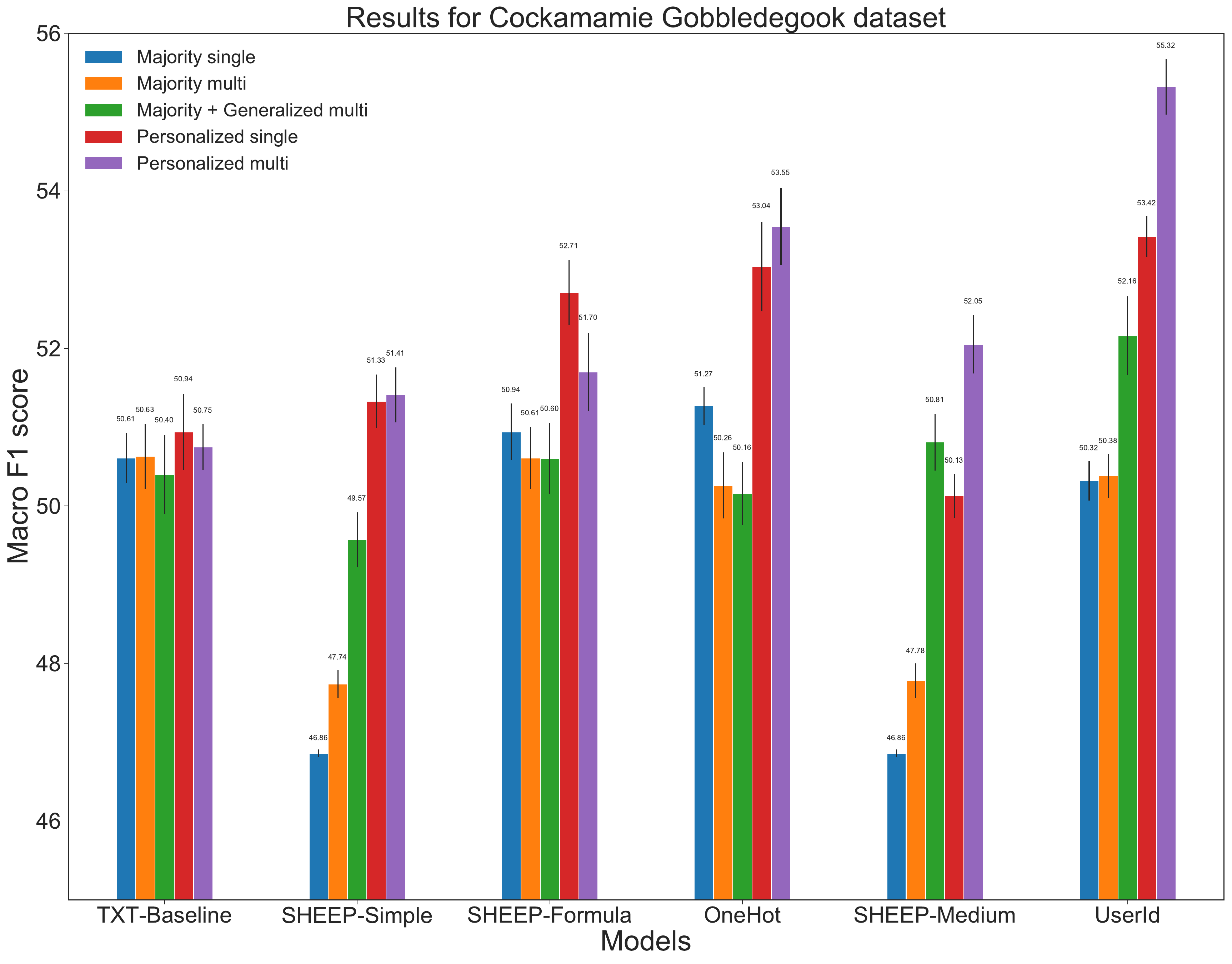}
    \caption{Macro F1 score values for all five experimental configurations tested on the personalized Cockamamie Gobbledegook dataset. The bar whiskers outline the standard deviation of the results measured during the cross-validation.}
    \label{fig:cockamamie_f1}
\end{figure}

To evaluate the impact of our human-centered data fusion techniques, we carried out experiments in configurations described in Sec.~\ref{sec:experimental_setup}. The results for the Cockamamie Gobbledegook dataset are presented in Tab.~\ref{fig:cockamamie_f1}. The best results were obtained using the UserId model. Almost every personalized architecture achieved the best results in Personalized multi configuration. The slightly lower results for the SHEEP-Formula model in this configuration in comparison to training on the original dataset may be related to the nature of this dataset containing texts no longer than 2 words. In addition, the words are very uncommon, making it even more difficult to reliably assess the uniqueness of the user's perception against others.

\begin{figure}[h]
    \centering
    \includegraphics[width=\linewidth]{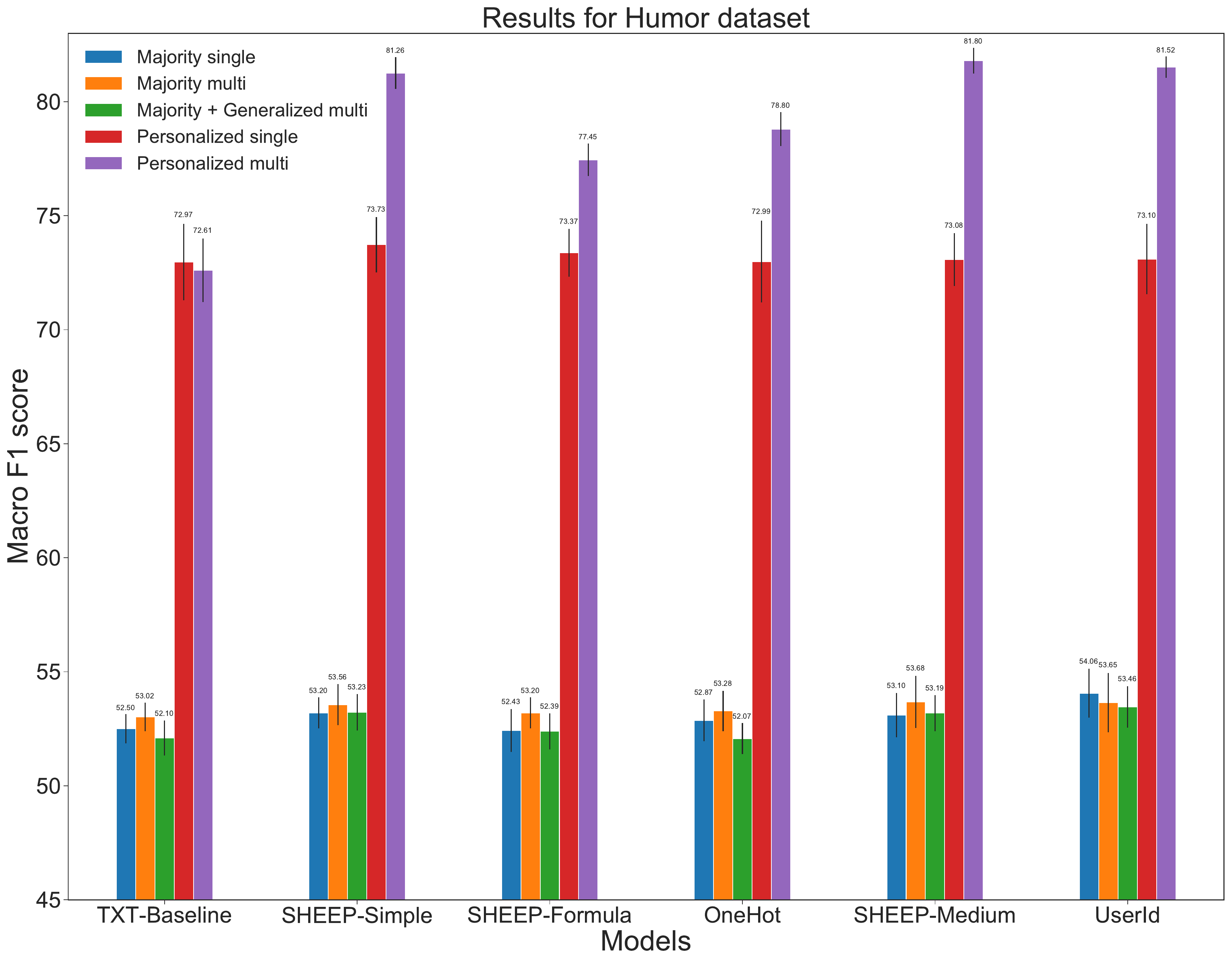}
    \caption{Macro F1 score values for all five experimental configurations tested on the personalized Humor dataset. The bar whiskers outline the standard deviation of the results measured during the cross-validation.}
    \label{fig:humor_f1}
\end{figure}

The evaluation results for the Humor dataset are presented in Fig.~\ref{fig:humor_f1}. The use of human-centered data fusion (Personalized multi) resulted in an improvement of up to 8.72 in macro F1 score compared to training on the original dataset (Personalized single). The best results were achieved by the SHEEP-Medium, UserId, and SHEEP-Simple models.

\begin{figure}[h]
    \centering
    \includegraphics[width=\linewidth]{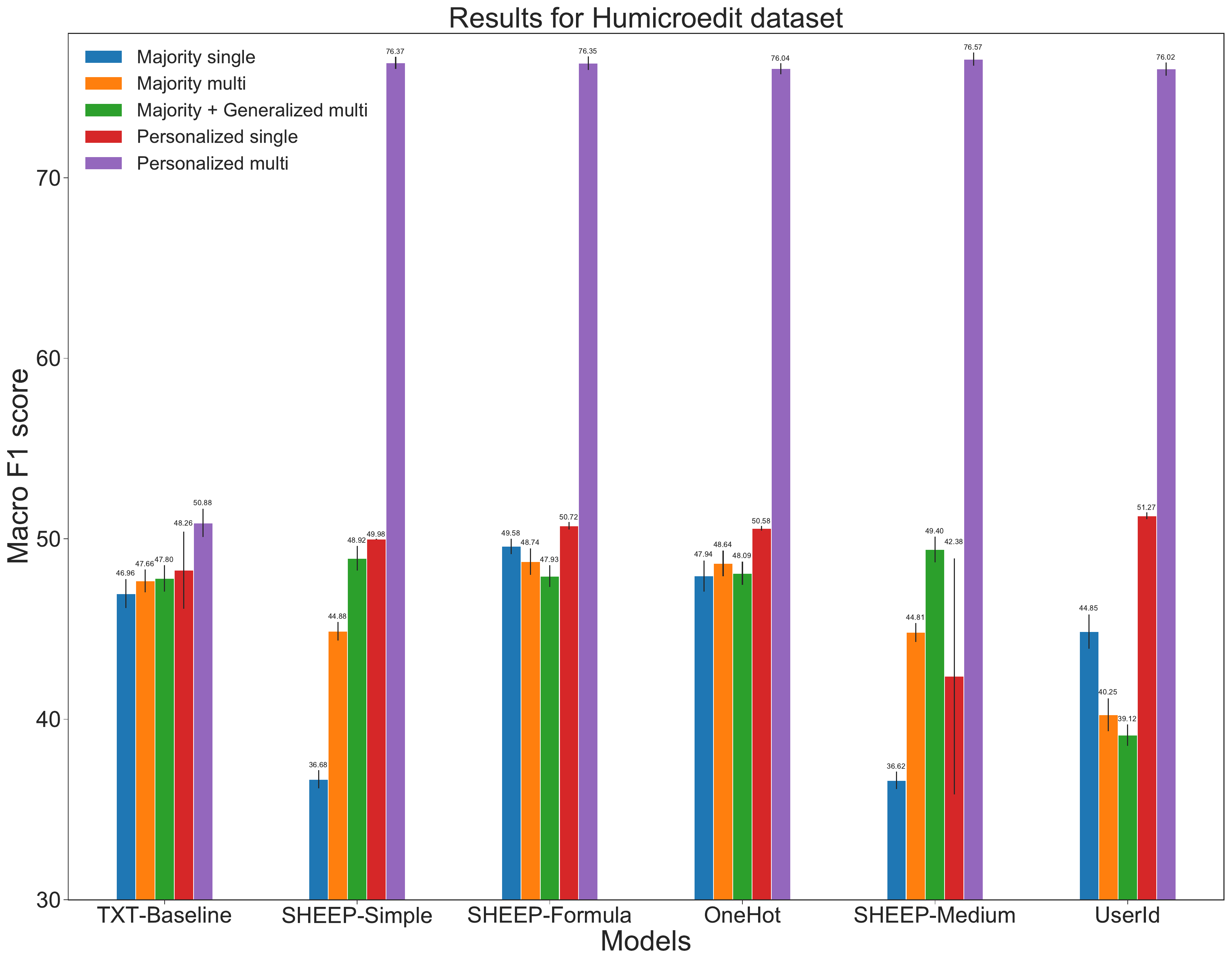}
    \caption{Macro F1 score values for all five experimental configurations tested on the personalized Humicroedit dataset. The bar whiskers outline the standard deviation of the results measured during the cross-validation.}
    \label{fig:humicroedit_f1}
\end{figure}

The results of the conducted on the Humicroedit dataset are shown in Fig.~\ref{fig:humicroedit_f1}. Every tested architecture achieved the best results by training on the combined personalized datasets with individual user annotations (Personalized multi). The highest difference in macro F1 score compared to training on the original dataset (Personalized single) is 34.19 for SHEEP-Medium. The significantly higher standard deviation of the results of this model observed in the personalized single configuration may be due to the fact that there are only five users in the entire dataset. Having nearly 15,000 annotations for each user makes it difficult for this model to generate a vector representation that would highlight the differences between the preferences of only five users for such a large number of annotations per user. Moreover, the stabilization of this standard deviation in the Personalized multi configuration is an additional benefit of applying our personalized data fusion.

\begin{figure}[h]
    \centering
    \includegraphics[width=\linewidth]{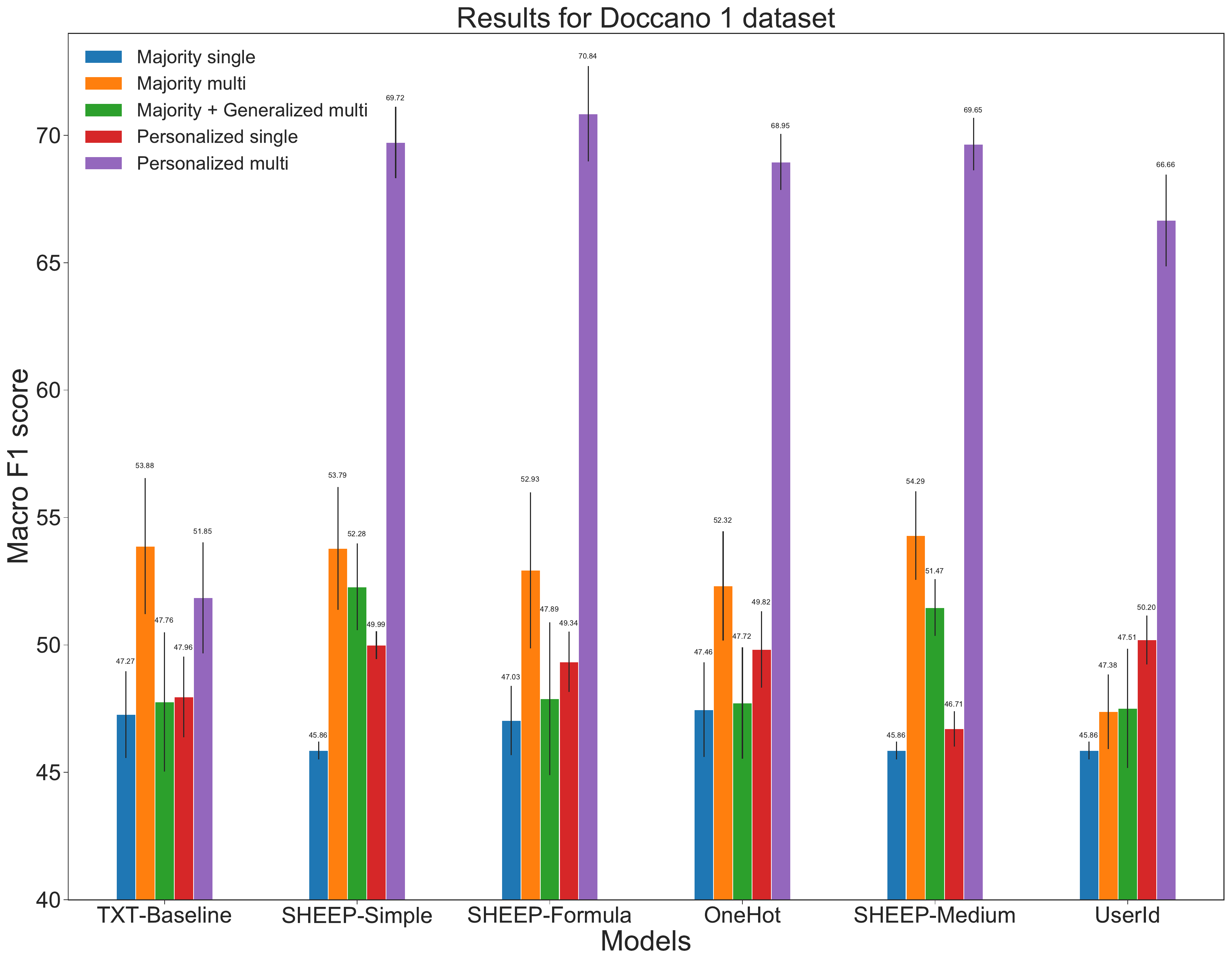}
    \caption{Macro F1 score values for all five experimental configurations tested on the personalized Docano 1 dataset. The bar whiskers outline the standard deviation of the results measured during the cross-validation.}
    \label{fig:doccano1_f1}
\end{figure}

Fig.~\ref{fig:doccano1_f1} shows the results on the Doccano 1 dataset. The best results were obtained in Personalized multi configuration for all personalized models. The use of our personalized data fusion technique allowed for achieving up to 21.50 improvement in macro F1 score compared to training on the original dataset (Personalized single). Furthermore, the use of our data fusion technique based on the personalized datasets aggregated by majority voting (Majority multi) resulted in significant performance improvements in comparison to training on the original dataset (Personalized single) for every model except UserId. This is caused by the use of a special user token in this architecture. While this mechanism helps the model to better obtain the user representation, in majority voting-based configurations, the same user token added at the start of a text sample causes only unnecessary noise that can obscure more important text features from the model or steer the learning process in the wrong direction.

\begin{figure}[h]
    \centering
    \includegraphics[width=\linewidth]{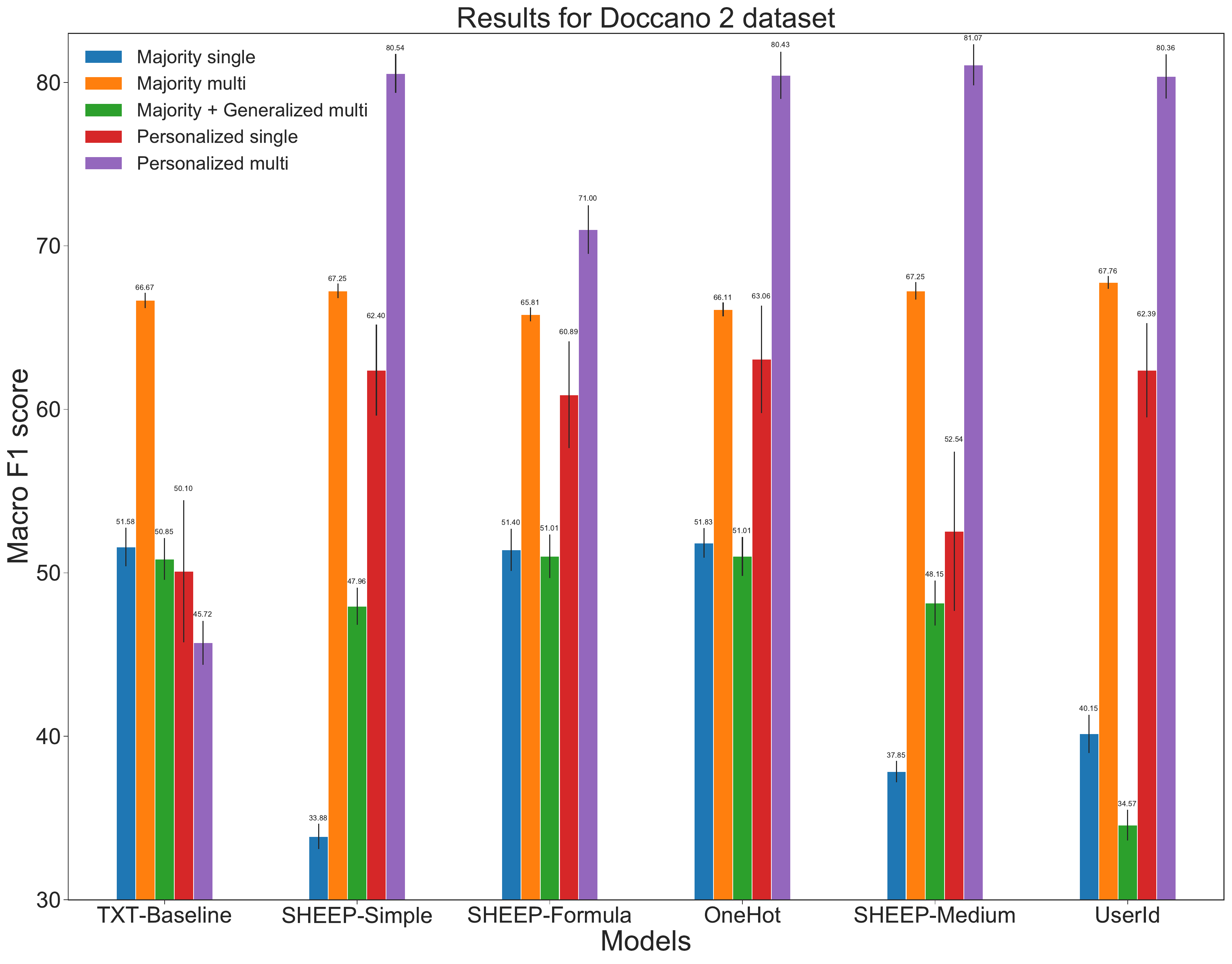}
    \caption{Macro F1 score values for all five experimental configurations tested on the personalized Doccano 2 dataset. The bar whiskers outline the standard deviation of the results measured during the cross-validation.}
    \label{fig:doccano2_f1}
\end{figure}

The results for the Doccano 2 dataset are shown in Fig.~\ref{fig:doccano2_f1}. The use of human-centered data fusion techniques (Personalized multi) resulted in achieving significant improvement in performance for all personalized architectures. The highest increase of 28.53 in macro F1 score was observed compared to training on the original data (Personalized single) for the SHEEP-Medium model. Moreover, every architecture (including the non-personalized TXT-Baseline model) achieved significantly better results by training on the combined personalized datasets aggregated via majority voting (Majority multi) in comparison to the results of a model trained on the original dataset (Personalized single). The lower performance of the TXT-Baseline in the Personalized multi configuration is related to the lack of leveraging any user information in this model. By ignoring any information about the user, the model sees the larger amount of diverse annotations as unnecessary noise, concealing the important patterns in the data. On the other hand, training the TXT-Baseline model on combined personalized datasets aggregated by majority voting (Majority multi) distilled the knowledge extractable by the model from the dataset and thus highlighted the most important patterns in the data.

\subsection{Knowledge Transfer in Majority Voting and Personalization}
\label{sec:knowledge_transfer_majority_personalization}

\begin{figure}[h]
    \centering
    \includegraphics[width=\linewidth]{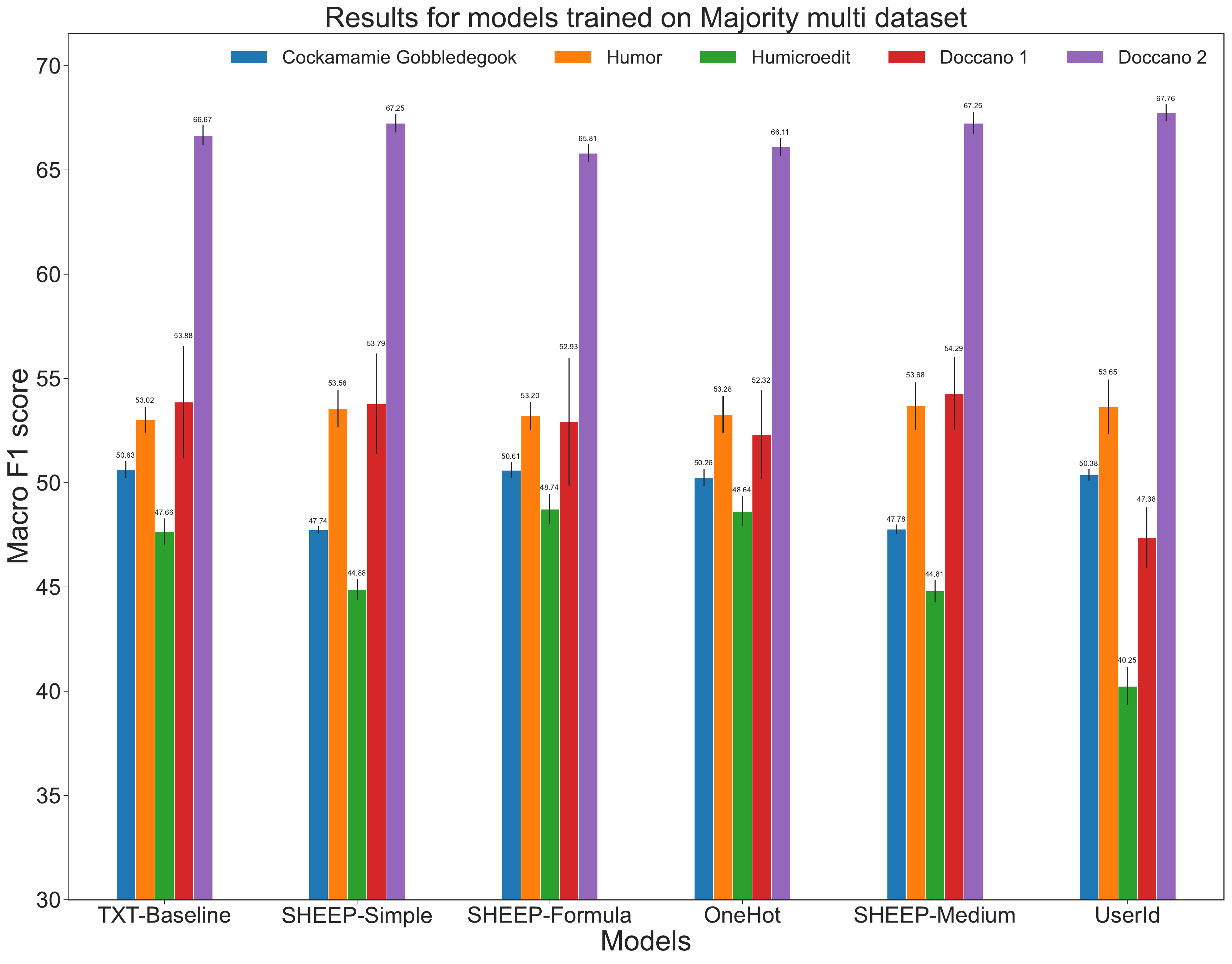}
    \caption{Macro F1 score values for all personalized datasets achieved by models trained on combined personalized datasets aggregated via majority voting (Majority multi). The bar whiskers outline the standard deviation of the results measured during the cross-validation.}
    \label{fig:merged_majority_f1}
\end{figure}

\begin{figure}[h]
    \centering
    \includegraphics[width=\linewidth]{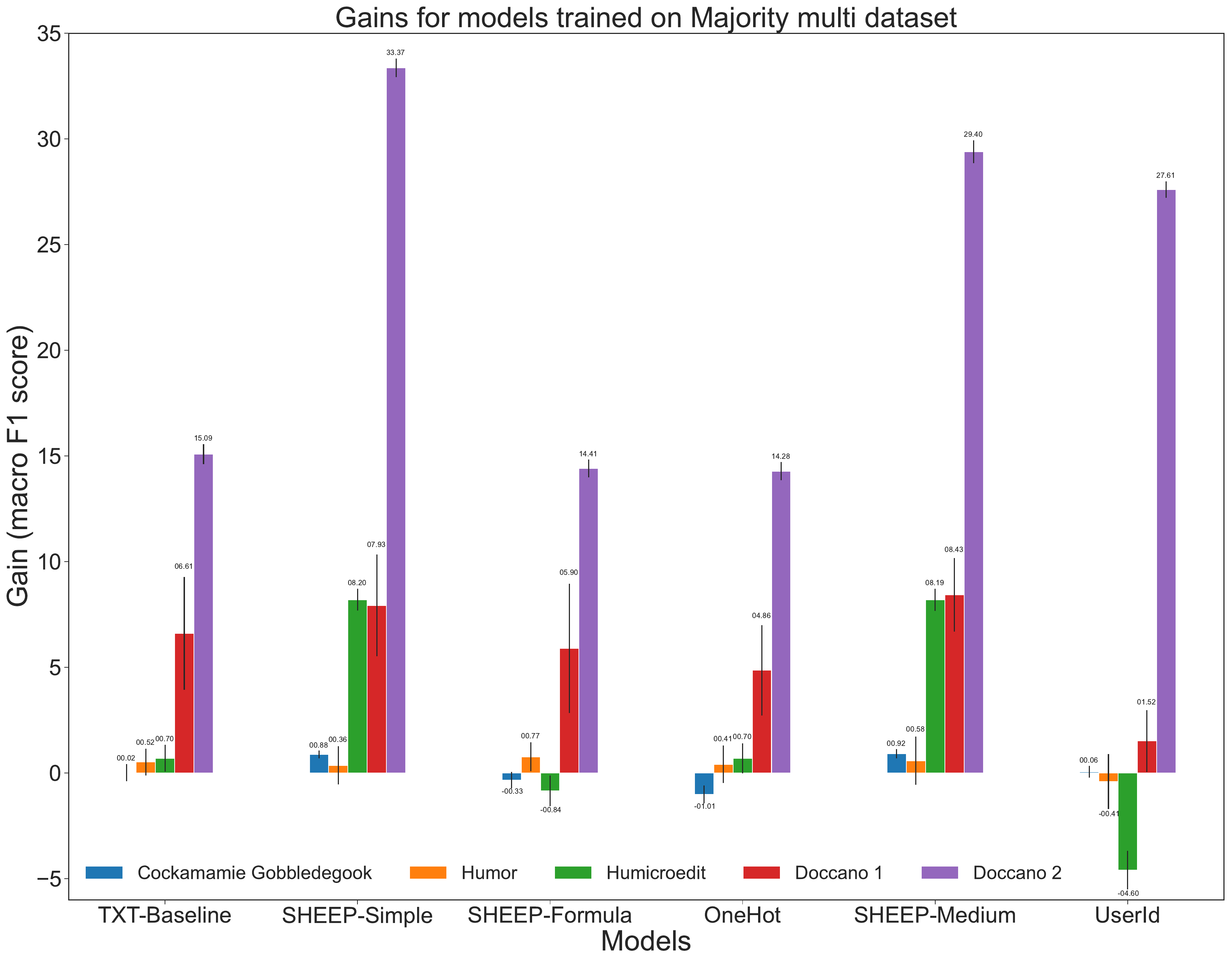}
    \caption{Difference between macro F1 score values obtained by models trained on combined personalized datasets aggregated via majority voting (Majority multi) and the models trained on a single personalized dataset aggregated via majority voting. The bar whiskers outline the standard deviation of the results measured during the cross-validation.}
    \label{fig:merged_majority_f1_gain}
\end{figure}

The evaluation results of models trained on the combined personalized datasets containing samples aggregated by majority voting (Majority multi) are presented in Fig.~\ref{fig:merged_majority_f1}. Similar results obtained by the TXT-Baseline model and all personalized architectures are caused by the lack of user information in this setup. On the other hand, the gains obtained via leveraging the majority voting-based variant of our data fusion technique are presented in Fig.~\ref{fig:merged_majority_f1_gain}. The values are equal to the difference between the macro F1 score obtained by the model trained on the combined personalized datasets aggregated by majority voting (Majority multi) and the model trained on the original data (Personalized single). The highest gains were observed for the Doccano 2 dataset. This may be related to the small size of this dataset, which, combined with very diverse assessments of a large number of users, may significantly hinder the model extraction of the knowledge needed in the humor recognition task from the representation of the text annotated in a very ambiguous way. The negative gains observed for the Humicroedit dataset in the case of the UserId model may be related to a much different profile of other datasets. This dataset is characterized by a large number of annotations and a small number of users, which lowers the variety of evaluations of individual texts. Therefore, the inter-annotator agreement is significantly higher than in the case of other personalized datasets. This results in a situation where adding information about texts annotated more ambiguously by a larger number of users may not change the general patterns helpful for this particular dataset, but only add unnecessary noise during the training procedure. The lower complexity of the SHEEP-Simple and SHEEP-Medium models makes them more resistant to this phenomenon. The lack of a proper increase for the Cockamamie Gobbledegook dataset may be related to the very low length of each sample in this dataset and the occurrence of very unique words, which may not have proper vector representations. The Humor dataset is the only personalized dataset containing texts in Spanish, which may reduce the gain from annotations regarding texts in different languages.

\begin{figure}[h]
    \centering
    \includegraphics[width=\linewidth]{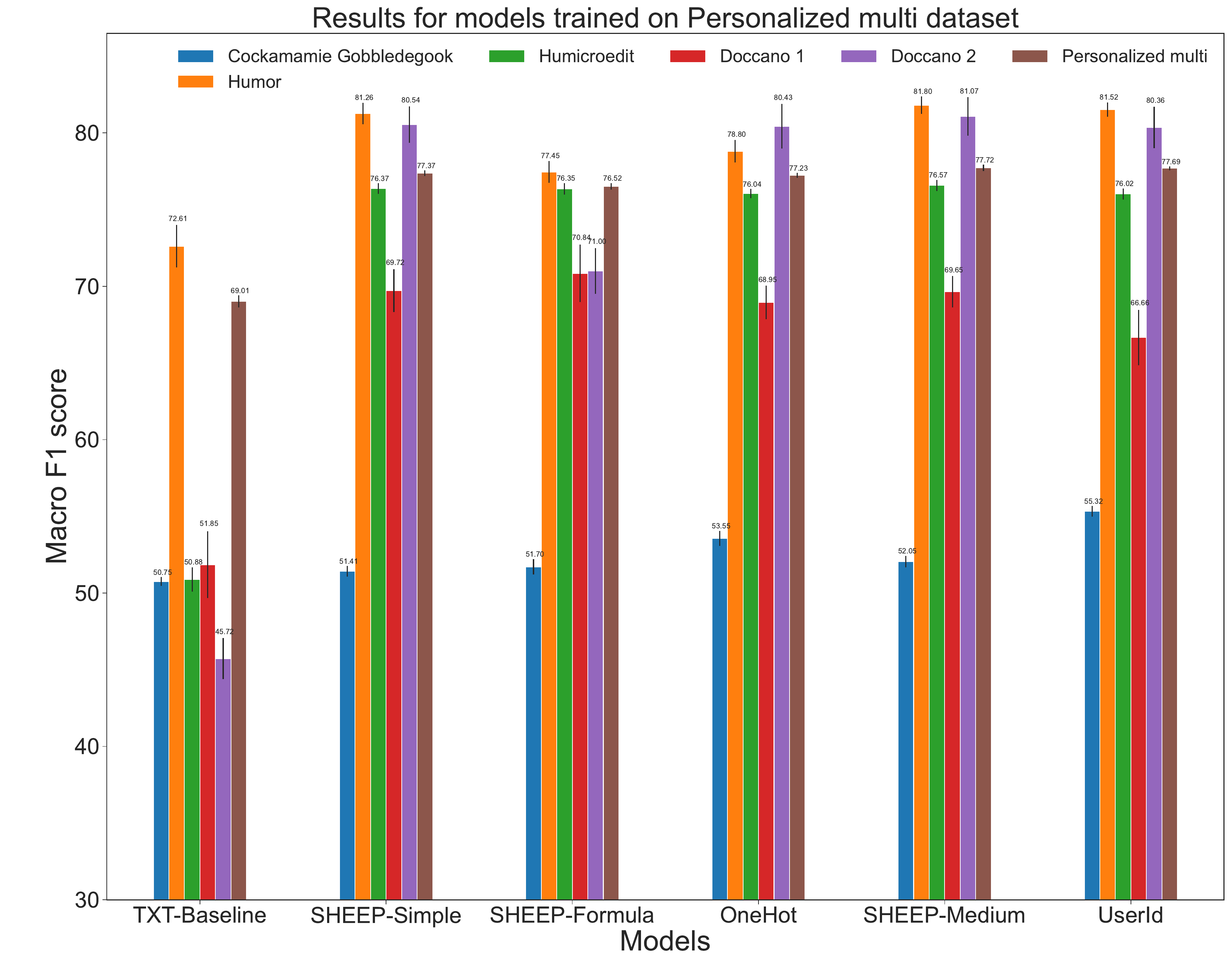}
    \caption{Macro F1 score values for all personalized datasets achieved by models trained on combined personalized datasets containing individual user annotations (Personalized multi). The bar whiskers outline the standard deviation of the results measured during the cross-validation.}
    \label{fig:merged_personalized_f1}
\end{figure}

\begin{figure}[h]
    \centering
    \includegraphics[width=\linewidth]{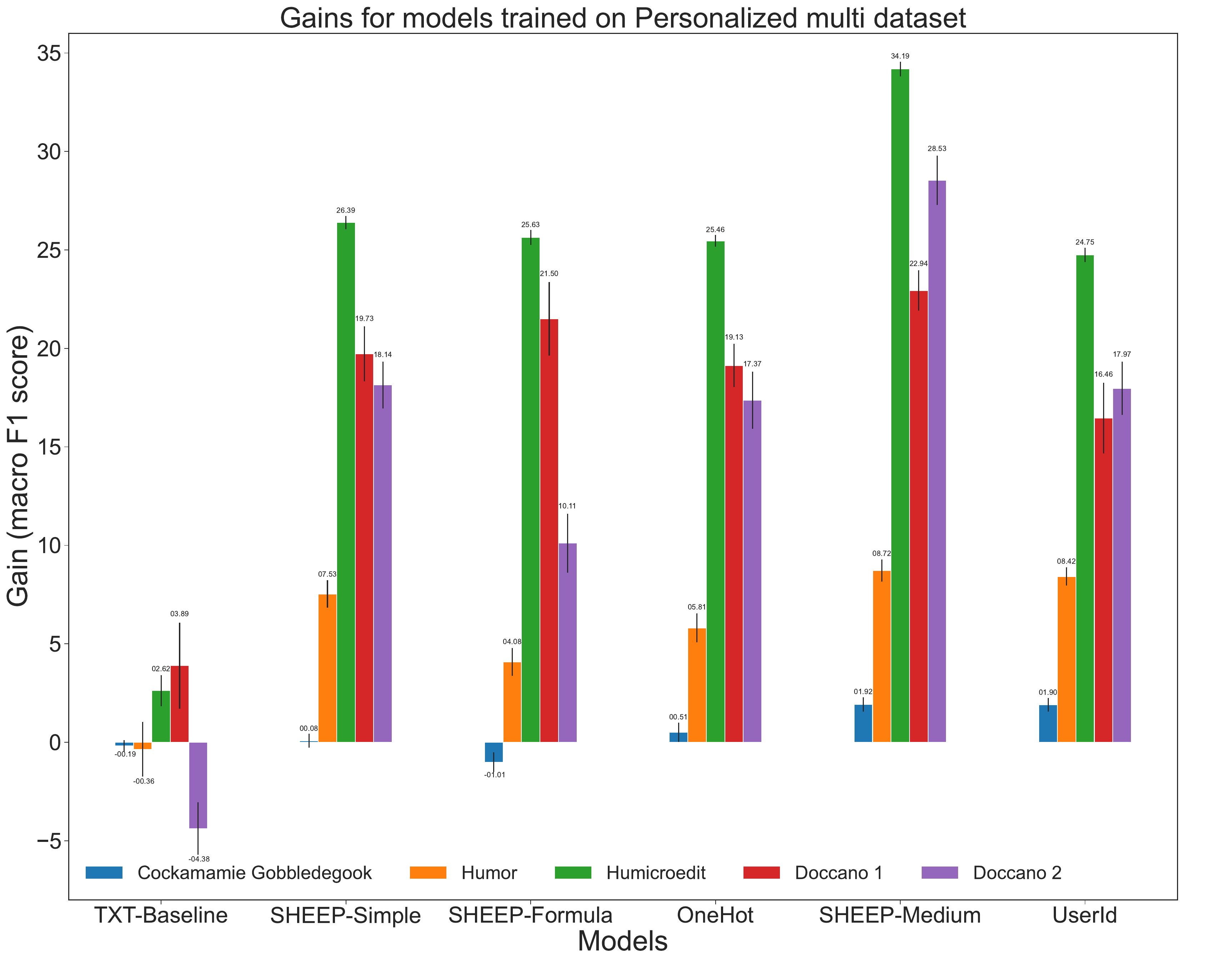}
    \caption{Difference between macro F1 score values obtained by models trained on combined personalized datasets with individual user annotations (Personalized multi) and the models trained on a single personalized dataset containing individual user annotations. The bar whiskers outline the standard deviation of the results measured during the cross-validation.}
    \label{fig:merged_personalized_f1_gain}
\end{figure}

The results of experiments involving models trained on fused personalized datasets with individual user annotations (Personalized multi) are presented in Fig.~\ref{fig:merged_personalized_f1}. The best results were achieved by the SHEEP-Formula, SHEEP-Medium, and UserId models. The gains obtained by the use of the proposed human-centered data fusion are shown in Fig.~\ref{fig:merged_personalized_f1_gain}. The values are equal to the difference in macro F1 score between the model trained on the combined personalized datasets with individual user annotations (Personalized multi) and the model trained on the original dataset (Personalized single). The highest gains are observed for the Humicroedit, Doccano 2, and Doccano 1 datasets. The first one originally lacked useful information to distinguish the user from the others. With the additional data from different datasets included during the training procedure, the model improved its ability to leverage information about the user and how to better focus on the differences between users. In the case of the Doccano 1 and Doccano 2 datasets, additional fused data contained not only information about other users but also allowed the model to extract general knowledge about the relation between the textual content and its funniness. The negative gain value for the Doccano 2 dataset in the case of the TXT-Baseline model may be related to the extension of the train set with texts in languages other than Polish. Although the language model used is language-agnostic, there is still a significant change in the distribution of embeddings in the train set, which could negatively affect the performance of the model, which does not use any information about the user. On the other hand, the lack of significant gains for the Cockamamie Gobbledegook dataset may be related to the nature of the samples present in this dataset. It contains texts no longer than two words and some of them are word-formations. It also interferes with the model's ability to generalize patterns extracted from other datasets to such specific samples.

\subsection{Impact of Data Fusion on Personalized Architectures}
\label{sec:data_fusion_on_architectures}

\begin{figure}[h]
    \centering
    \includegraphics[width=\linewidth]{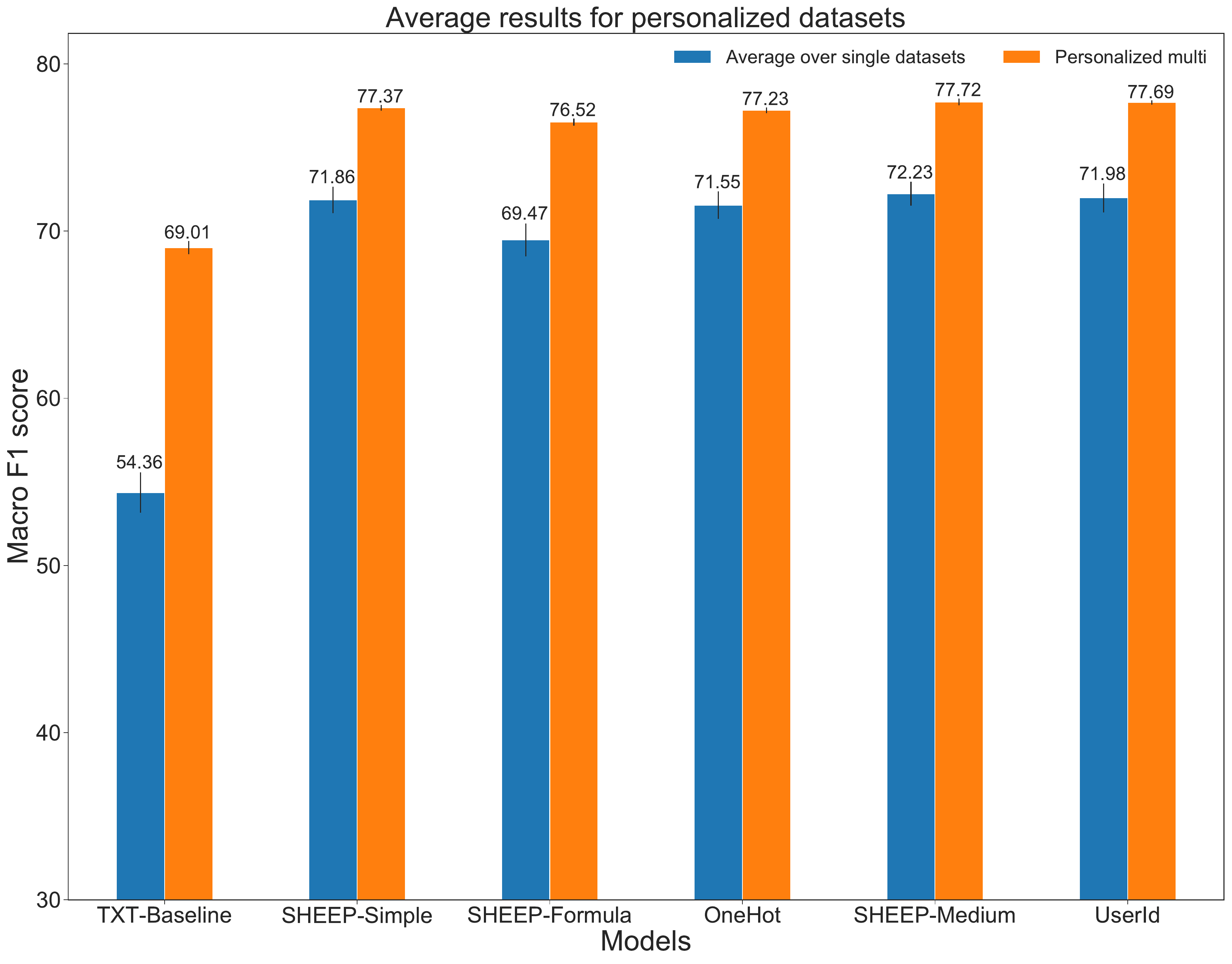}
    \caption{Average macro F1 scores achieved by the personalized architectures trained on single personalized datasets with individual user annotations compared with the performance of models trained on combined personalized datasets containing individual user annotaions (Personalized multi).}
    \label{fig:avg_model_results}
\end{figure}

When comparing the effectiveness of our human-centered data fusion techniques in TXT-Baseline and the personalized architectures presented in Fig.~\ref{fig:avg_model_results}, it is prominent that each individual case favors training on the combined personalized datasets (Personalized multi) over training on the single personalized datasets (Personalized single). This outlines the versatile nature of the proposed technique and the possibility of applying it regardless of the chosen model architecture. The influence of combining many different personalized datasets in this matter indicates that due to the subjective nature of humor perception, better results may be obtained by increasing the variety of user perspectives seen by the model during the training procedure. A higher diversity of annotations can also improve the quality of the patterns extracted by the model from the data, increasing their consistency with the real characteristics of the phenomenon, as indicated by the improvement of the TXT-Baseline scores.

\section{Discussion}
\label{sec:discussion}
Experimental studies in nine datasets, of which the majority is personalized, revealed that the concatenation of all subjective data massively increases the reasoning quality of a model, which is especially prominent in Fig.~\ref{fig:humicroedit_f1} and Fig.~\ref{fig:doccano2_f1}.

The smaller gains observed in the Cockamamie Gobbledegook dataset in Fig.~\ref{fig:cockamamie_f1} are the result of the word-formation characteristics of this data collection, such as where headlines (Humicroedit) and tweets (Humor) contain a more standard form in which humor is contained. 

Another crucial aspect of the results is the fact that the language of the dataset also affected the level of increase in reasoning performance. The general results of the experiments on the datasets are also linked to the language of its content. There is no doubt that certain features related to English, Spanish, and Polish are involved in the research, although a language-agnostic model is used to generate text representations. As seen in Fig.~\ref{fig:merged_personalized_f1_gain}, the model performance on the Spanish dataset Humor is noticeably worse than on the English and Polish data, as it may be the result of concatenating multiple different languages with vastly different characteristics.

Through our research, each of our experimental results shows that no matter how much we increase the quantity of a generalized dataset in the Generalized + Majority scenario, it will never exceed the boost guaranteed by the incorporation of combined personalized datasets in the Personalized multi scenario, as seen in Fig.~\ref{fig:avg_model_results}. This fact strongly highlights the subjective nature of the humor recognition task in NLP. The message indicated by the results of our experiments implies that knowledge about the user's beliefs, feelings, and experiences is much more crucial than information regarding the text itself. The incorporation of generalized data sets did not improve the quality of reasoning, regardless of the selected model. This could be caused by the fact that the annotation process in the case of generalized datasets was focused on maximizing the inter-annotator agreement. As a result, this approach discards the information about the funniness of the text that includes different perspectives of the anotators. It is evident that if an annotation process is to be performed, it should take into account the individual perspective of the user.

As for our majority-based data fusion method (Majority multi) based on the majority voting in personalized datasets, it can also be used to improve the performance of the non-personalized models as can be observed for the TXT-Baseline model in Fig.~\ref{fig:avg_model_results}. 

On the other hand, similar results of TXT-Baseline and personalized architectures trained on datasets aggregated by majority voting (Majority multi) seen in Fig.~\ref{fig:merged_majority_f1} indicate the universal nature of the personalized architectures used in the experiments and the possibility of their application to non-personalized datasets. This could allow for a synthesis of knowledge from the humor domain from multiple sources, which enhances the overall versatility of the model performance.



\section{Conclusions and Future work}
\label{sec:conclusions_future_work}
Our research experiments allowed us to recognize the importance of human-centered data fusion for natural language processing reasoning, especially in terms of work in the area of subjective NLP. Contrary to generalized NLP problems, here it is only the user who decides what is funny or not, and this fundamental fact is the key factor in unlocking the potential of personalized reasoning. We have proven that even if the amount of data in generalized scenarios is much greater, the aspect of user preference appears to be the best out of all five scenarios considered.

We have demonstrated that incorporating knowledge from other datasets can be beneficial for the model by improving its general understanding of the task of humor detection. The fusion of data regarding an individual user's sense of humor from multiple sources has an effect on the model performance on other personalized datasets in both personalized and majority-voting-based situations. The incorporation of generalized datasets improved the effectiveness of humor prediction for personalized datasets in a majority voting context, and knowledge transfer between personalized datasets has a similar effect on model performance as knowledge transfer between datasets with majority voting.

The results of data fusion were found to depend on the language or field of the datasets, and it was concluded that the use of human-centered data fusion techniques improves the performance of a variety of personalized structures in a similar way.

We are confident that our method of data fusion is capable of accurately representing variety of user beliefs and this could be the way forward for subjective NLP tasks, such as humor detection and any other. Therefore, we plan to adapt our human-centered data fusion techniques to other subjective NLP tasks. In our other future research, we want to exploit categories of humor in a completely personalized dataset that could combine the humor perception of each user. Every individual may have a distinct sense of humor, so it is essential to recognize any potential feature where the perception and humor standard may vary. Such identification would enable us to compare people more precisely, thus uncovering the element of a potential individualized rating system. 

The source code used during the experiments is publicly available \footnote{\url{https://github.com/CLARIN-PL/personalized-nlp/releases/tag/2023-icdm-sentire-humor}}.

\section*{Acknowledgements}
This work was financed by 
(1) the National Science Centre, Poland, project no. 2021/41/B/ST6/04471;  
(2) Contribution to the European Research Infrastructure 'CLARIN ERIC - European Research Infrastructure Consortium: Common Language Resources and Technology Infrastructure', 2022-23 (CLARIN Q);
(3) the Polish Ministry of Education and Science, CLARIN-PL; 
(4) the European Regional Development Fund as a part of the 2014-2020 Smart Growth Operational Programme, projects no. POIR.04.02.00-00C002/19, POIR.01.01.01-00-0288/22 and POIR.01.01.01-00-0923/20; 
(5) the statutory funds of the Department of Artificial Intelligence, Wroclaw University of Science and Technology;
(6) the Polish Ministry of Education and Science within the programme “International Projects Co-Funded”;
(7) the European Union under the Horizon Europe, grant no. 101086321 (OMINO). However, the views and opinions expressed are those of the author(s) only and do not necessarily reflect those of the European Union or the European Research Executive Agency. Neither the European Union nor European Research Executive Agency can be held responsible for them.

\bibliographystyle{IEEEtran}
\bibliography{main}

\begin{thebibliography}{10}
\providecommand{\url}[1]{#1}
\csname url@samestyle\endcsname
\providecommand{\newblock}{\relax}
\providecommand{\bibinfo}[2]{#2}
\providecommand{\BIBentrySTDinterwordspacing}{\spaceskip=0pt\relax}
\providecommand{\BIBentryALTinterwordstretchfactor}{4}
\providecommand{\BIBentryALTinterwordspacing}{\spaceskip=\fontdimen2\font plus
\BIBentryALTinterwordstretchfactor\fontdimen3\font minus
  \fontdimen4\font\relax}
\providecommand{\BIBforeignlanguage}[2]{{%
\expandafter\ifx\csname l@#1\endcsname\relax
\typeout{** WARNING: IEEEtran.bst: No hyphenation pattern has been}%
\typeout{** loaded for the language `#1'. Using the pattern for}%
\typeout{** the default language instead.}%
\else
\language=\csname l@#1\endcsname
\fi
#2}}
\providecommand{\BIBdecl}{\relax}
\BIBdecl

\bibitem{kanclerz2020cross}
K.~Kanclerz \emph{et~al.}, ``Cross-lingual deep neural transfer learning in
  sentiment analysis,'' \emph{Procedia Computer Science}, vol. 176, pp.
  128--137, 2020.

\bibitem{kanclerz2021controversy}
------, ``Controversy and conformity: from generalized to personalized
  aggressiveness detection,'' in \emph{Proceedings of the 59th Annual Meeting
  of the Association for Computational Linguistics and the 11th International
  Joint Conference on Natural Language Processing (Volume 1: Long Papers)},
  2021, pp. 5915--5926.

\bibitem{hofmann2020gender}
J.~Hofmann \emph{et~al.}, ``Gender differences in humor-related traits, humor
  appreciation, production, comprehension,(neural) responses, use, and
  correlates: A systematic review,'' \emph{Current Psychology}, pp. 1--14,
  2020.

\bibitem{kocon2021learning}
J.~Koco{\'n} \emph{et~al.}, ``Learning personal human biases and
  representations for subjective tasks in natural language processing,'' in
  \emph{2021 IEEE International Conference on Data Mining (ICDM)}.\hskip 1em
  plus 0.5em minus 0.4em\relax IEEE, 2021, pp. 1168--1173.

\bibitem{kocon2023chatgpt}
------, ``Chatgpt: Jack of all trades, master of none,'' \emph{Information
  Fusion}, p. 101861, 2023.

\bibitem{mieleszczenko2023capturing}
W.~Mieleszczenko-Kowszewicz \emph{et~al.}, ``Capturing human perspectives in
  nlp: Questionnaires, annotations, and biases,'' 2023.

\bibitem{kazienko2023human}
P.~Kazienko, J.~Bielaniewicz, M.~Gruza, K.~Kanclerz, K.~Karanowski,
  P.~Mi{\l}kowski, and J.~Koco{\'n}, ``Human-centered neural reasoning for
  subjective content processing: Hate speech, emotions, and humor,''
  \emph{Information Fusion}, vol.~94, pp. 43--65, 2023.

\bibitem{wilson1969conservatism}
G.~D. Wilson and J.~R. Patterson, ``Conservatism as a predictor of humor
  preferences.'' \emph{Journal of consulting and clinical Psychology}, p. 271,
  1969.

\bibitem{bekelja2006appropriate}
B.~W. others, ``Appropriate and inappropriate uses of humor by teachers,''
  \emph{Communication education}, pp. 178--196, 2006.

\bibitem{hay1995gender}
J.~Hay, ``Gender and humour: Beyond a joke,'' \emph{Unpublished Master’s
  thesis, Victoria University of Wellington, Wellington, New Zealand}, 1995.

\bibitem{strapparava2011computational}
C.~Strapparava \emph{et~al.}, ``Computational humour,'' \emph{Emotion-Oriented
  Systems: The Humaine Handbook}, pp. 609--634, 2011.

\bibitem{raz2012automatic}
Y.~Raz, ``Automatic humor classification on twitter,'' in \emph{Proceedings of
  the NAACL HLT 2012 student research workshop}, 2012, pp. 66--70.

\bibitem{khandelwal2018humor}
A.~Khandelwal \emph{et~al.}, ``Humor detection in english-hindi code-mixed
  social media content: Corpus and baseline system,'' \emph{arXiv preprint
  arXiv:1806.05513}, 2018.

\bibitem{dynel2009beyond}
M.~Dynel, ``Beyond a joke: Types of conversational humour,'' \emph{Language and
  linguistics compass}, vol.~3, no.~5, pp. 1284--1299, 2009.

\bibitem{davies2017sociolinguistic}
C.~E. Davies, ``Sociolinguistic approaches to humor,'' \emph{The Routledge
  handbook of language and humor}, vol.~1, 2017.

\bibitem{agrawal2020joke}
S.~Agrawal, ``Joke recommender system using humor theory,'' Ph.D. dissertation,
  Purdue University, 2020.

\bibitem{meghana2020humour}
J.~Meghana and R.~Vijaya, ``Humour and gender stereotypes.'' \emph{IASSI
  Quarterly}, 2020.

\bibitem{siddiqui2019sarcasmania}
R.~Siddiqui, ``Sarcasmania: Sarcasm exposed,'' 2019.

\bibitem{bleiholder2009data}
J.~Bleiholder and F.~Naumann, ``Data fusion,'' \emph{ACM computing surveys
  (CSUR)}, pp. 1--41, 2009.

\bibitem{xu2022hybrid}
H.~Xu \emph{et~al.}, ``Hybrid multimodal fusion for humor detection,'' in
  \emph{Proceedings of the 3rd International on Multimodal Sentiment Analysis
  Workshop and Challenge}, 2022, pp. 15--21.

\bibitem{christ2022multimodal}
L.~Christ \emph{et~al.}, ``Multimodal prediction of spontaneous humour: A novel
  dataset and first results,'' \emph{arXiv preprint arXiv:2209.14272}, 2022.

\bibitem{christ2023muse}
------, ``The muse 2023 multimodal sentiment analysis challenge: Mimicked
  emotions, cross-cultural humour, and personalisation,'' \emph{arXiv preprint
  arXiv:2305.03369}, 2023.

\bibitem{yang2015}
D.~Yang \emph{et~al.}, ``Humor recognition and humor anchor extraction,'' in
  \emph{Proceedings of the 2015 Conference on Empirical Methods in Natural
  Language Processing}.\hskip 1em plus 0.5em minus 0.4em\relax Lisbon,
  Portugal: Association for Computational Linguistics, Sep. 2015, pp.
  2367--2376.

\bibitem{castro2018}
S.~Castro \emph{et~al.}, ``A crowd-annotated spanish corpus for humor
  analysis,'' in \emph{Proc. of SocialNLP 2018}.\hskip 1em plus 0.5em minus
  0.4em\relax ACL, 2018, pp. 7--11.

\bibitem{humicroedit_dataset}
N.~Hossain \emph{et~al.}, ``{``}president vows to cut
  {\textless}taxes{\textgreater} hair{''}: Dataset and analysis of creative
  text editing for humorous headlines,'' in \emph{Proc. of the NAACL: Human
  Language Technologies}.\hskip 1em plus 0.5em minus 0.4em\relax ACL, 2019, pp.
  133--142.

\bibitem{hossain-etal-2020-semeval}
------, ``{S}em{E}val-2020 task 7: Assessing humor in edited news headlines,''
  in \emph{SemEval2020}.\hskip 1em plus 0.5em minus 0.4em\relax ICCL, 2020.

\bibitem{bielaniewicz2022}
J.~Bielaniewicz \emph{et~al.}, ``Deep-sheep: Sense of humor extraction from
  embeddings in the personalized context,'' in \emph{2022 IEEE International
  Conference on Data Mining Workshops (ICDMW)}, 2022, pp. 967--974.

\bibitem{ziv1979sociometry}
A.~Ziv, ``Sociometry of humor: Objectifying the subjective,'' \emph{Perceptual
  and Motor Skills}, pp. 97--98, 1979.

\bibitem{engelthaler2018humor}
T.~Engelthaler \emph{et~al.}, ``Humor norms for 4,997 english words,''
  \emph{Behavior research methods}, vol.~50, no.~3, pp. 1116--1124, 2018.

\bibitem{annamoradnejad2020colbert}
I.~Annamoradnejad and G.~Zoghi, ``Colbert: Using bert sentence embedding for
  humor detection,'' \emph{arXiv preprint arXiv:2004.12765}, 2020.

\bibitem{chiruzzo2021overview}
L.~Chiruzzo \emph{et~al.}, ``Overview of haha at iberlef 2021: Detecting,
  rating and analyzing humor in spanish,'' \emph{Procesamiento del Lenguaje
  Natural}, pp. 257--268, 2021.

\bibitem{chiruzzo2020haha}
------, ``Haha 2019 dataset: A corpus for humor analysis in spanish,'' in
  \emph{Proceedings of the Twelfth Language Resources and Evaluation
  Conference}, 2020, pp. 5106--5112.

\bibitem{meaney-etal-2021-semeval}
J.~A. Meaney \emph{et~al.}, ``{S}em{E}val 2021 task 7: {H}a{H}ackathon,
  detecting and rating humor and offense,'' in \emph{Proceedings of the 15th
  International Workshop on Semantic Evaluation (SemEval-2021)}.\hskip 1em plus
  0.5em minus 0.4em\relax Online: Association for Computational Linguistics,
  Aug. 2021, pp. 105--119.

\bibitem{kocon2021ipm}
J.~Kocoń \emph{et~al.}, ``Offensive, aggressive, and hate speech analysis:
  From data-centric to human-centered approach,'' \emph{Information Processing
  \& Management}, vol.~58, no.~5, p. 102643, 2021.

\bibitem{student1908}
Student, ``The probable error of a mean,'' \emph{Biometrika}, vol.~6, no.~1,
  pp. 1--25, 1908.

\bibitem{bonferroni1936teoria}
C.~Bonferroni, \emph{Teoria statistica delle classi e calcolo delle
  probabilit{\`a}}, ser. Pubblicazioni del R. Istituto superiore di scienze
  economiche e commerciali di Firenze.\hskip 1em plus 0.5em minus 0.4em\relax
  Seeber, 1936.

\bibitem{mann1947}
H.~B. Mann and D.~R. Whitney, ``On a test of whether one of two random
  variables is stochastically larger than the other,'' \emph{The Annals of
  Mathematical Statistics}, vol.~18, no.~1, pp. 50--60, 1947.

\bibitem{sokolova2006beyond}
M.~Sokolova \emph{et~al.}, ``Beyond accuracy, f-score and roc: a family of
  discriminant measures for performance evaluation,'' in \emph{Australasian
  joint conference on artificial intelligence}.\hskip 1em plus 0.5em minus
  0.4em\relax Springer, 2006, pp. 1015--1021.

\bibitem{feng-etal-2022-language}
F.~Feng \emph{et~al.}, ``Language-agnostic {BERT} sentence embedding,'' in
  \emph{Proceedings of the 60th Annual Meeting of the Association for
  Computational Linguistics (Volume 1: Long Papers)}.\hskip 1em plus 0.5em
  minus 0.4em\relax Dublin, Ireland: Association for Computational Linguistics,
  May 2022, pp. 878--891.

\bibitem{wolf-etal-2020-transformers}
T.~Wolf \emph{et~al.}, ``Transformers: State-of-the-art natural language
  processing,'' in \emph{Proceedings of the 2020 Conference on Empirical
  Methods in Natural Language Processing: System Demonstrations}.\hskip 1em
  plus 0.5em minus 0.4em\relax Online: Association for Computational
  Linguistics, Oct. 2020, pp. 38--45.

\end{thebibliography}


\end{document}